\documentclass[letterpaper, 10 pt, conference]{ieeeconf}

\IEEEoverridecommandlockouts                              

\usepackage{listings}
\usepackage{float}
\usepackage[dvipdfmx]{graphicx}
\usepackage{amssymb}
\usepackage{latexsym}
\usepackage{amsfonts}
\usepackage{url}
\usepackage{comment}
\usepackage{algorithm}
\usepackage{algpseudocode}

\author{Koji Takeda$^{1}$, Kanji Tanaka$^{2}$ and Yoshimasa Nakamura$^{1}$
\thanks{$^{1}$K.Takeda and Y.Nakamura are with Tokyo Metropolitan Industrial Technology Research Institute,
         Tokyo, Japan
        {\tt\small \{takeda.koji\_1, nakamura.yoshimasa\}@iri-tokyo.jp}}%
\thanks{$^{2}$K. Tanaka is with Faculty of Engineering, University of Fukui, Japan.
        {\tt\small tanakakanji@gmail.com}}%
}

\begin{document}

\newcommand{\subcaption}[1]{\footnotesize #1}

\newcommand{\figA}{
\begin{figure}[t]
  \centering
  \includegraphics[width=8cm]{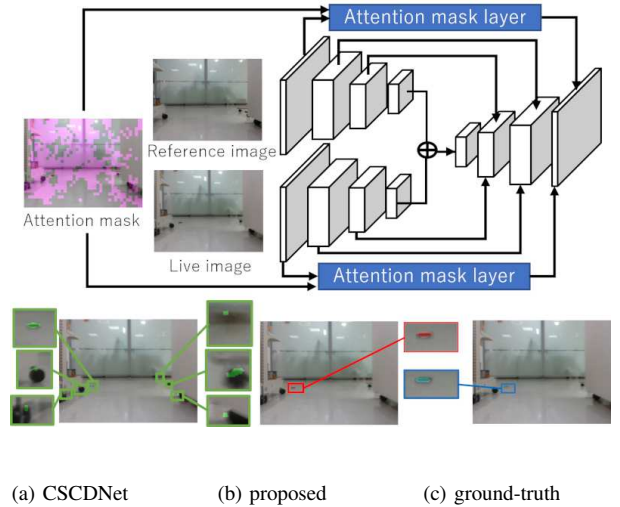}\vspace*{-3mm}\\
\begin{flushleft}
{\footnotesize
~~~~~~~~~ (a) CSCDNet 
~~~~~~~~~~ (b) proposed 
~~~~~~~~~~~ (c) ground-truth
}
\vspace*{-3mm}
\end{flushleft}
\caption{%
We present a novel attention technique
with 
an ability of 
unsupervised
on-the-fly domain adaptation,
which can
boost 
a state-of-the-art image change detection model,
by introducing an attention mask into the intermediate layer 
of the model,
without modifying the input and output layers.
}
\label{fig:tobirae}
\end{figure}
}

\newcommand{\figB}{
\hspace{20mm}
\begin{figure*}[t]
  \centering
  \includegraphics[width=17cm]{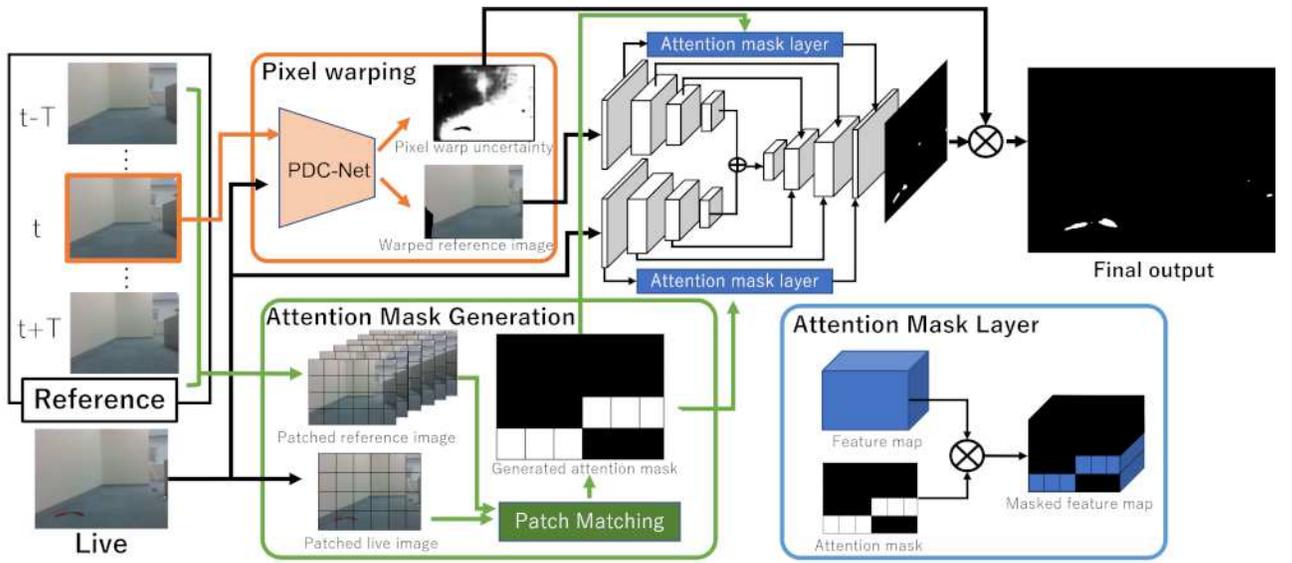}
  \caption{Image change detection framework.}
\label{fig:full}
\end{figure*}
}

\newcommand{\figC}{
\hspace{20mm}
\begin{figure}[t]
\centering
\includegraphics[width=5cm]{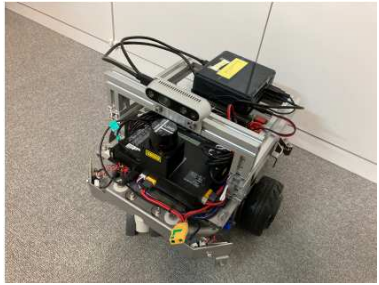}
\caption{Indoor robot experimental platform.}\label{fig:mobile_robot}
\end{figure}
}

\newcommand{\figD}{
\begin{figure}[t]
  \begin{minipage}[b]{2.8cm}
    \centering
    \includegraphics[width=2.8cm]{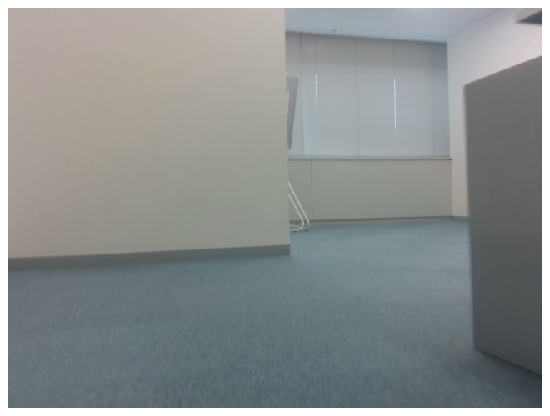}
    \subcaption{(a) Reference image}
  \end{minipage}
  \begin{minipage}[b]{2.8cm}
    \centering
    \includegraphics[width=2.8cm]{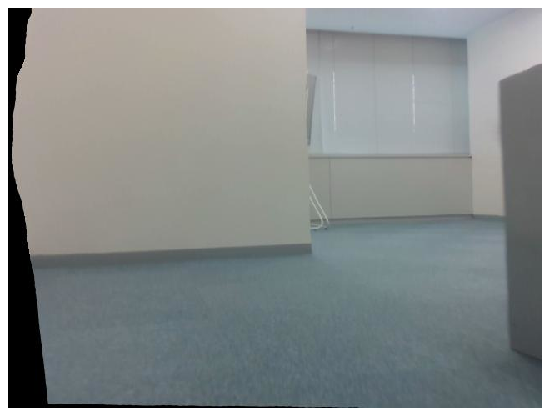}
    \subcaption{(b) Warped image}
  \end{minipage}
  \begin{minipage}[b]{2.8cm}
    \centering
    \includegraphics[width=2.8cm]{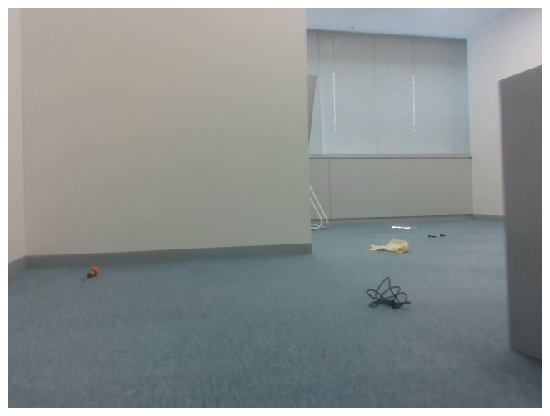}
    \subcaption{(c) Live image}
  \end{minipage}
  \caption{Example of pixel warping.}
  \label{fig:pixel_warp}
\end{figure}
}

\newcommand{\figE}{
\begin{figure}[t]
\includegraphics[width=8cm]{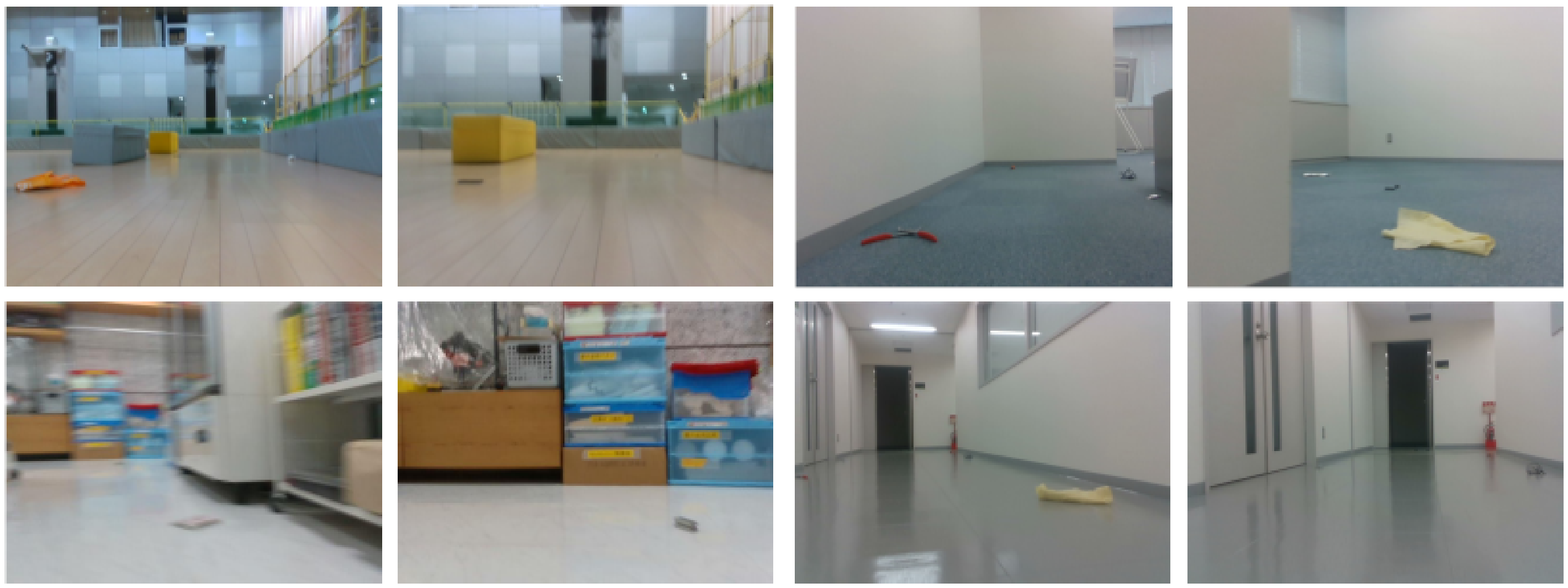}
\caption{Small object change detection dataset.}
\label{fig:dataset}
\end{figure}
}

\newcommand{\figF}{
\begin{figure}[t]
  \begin{minipage}[b]{2.7cm}
    \centering
    \includegraphics[width=2.6cm]{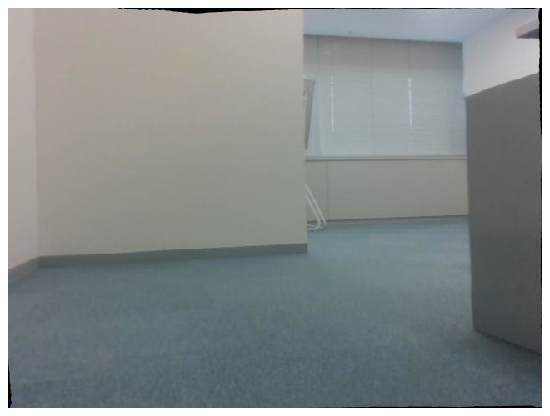}
    \subcaption{(a) Reference image}
  \end{minipage}
  \begin{minipage}[b]{2.7cm}
    \centering
    \includegraphics[width=2.6cm]{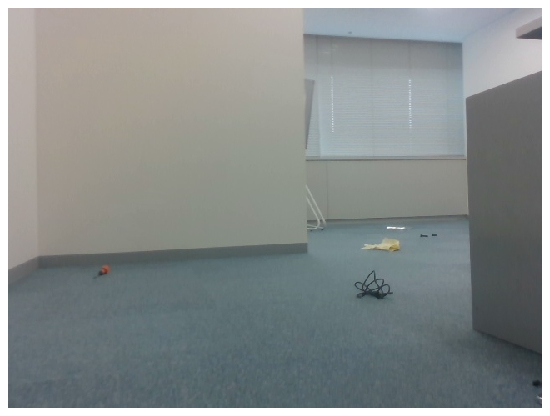}
    \subcaption{(b) Live image}
  \end{minipage}
  \begin{minipage}[b]{2.7cm}
    \centering
    \includegraphics[width=2.6cm]{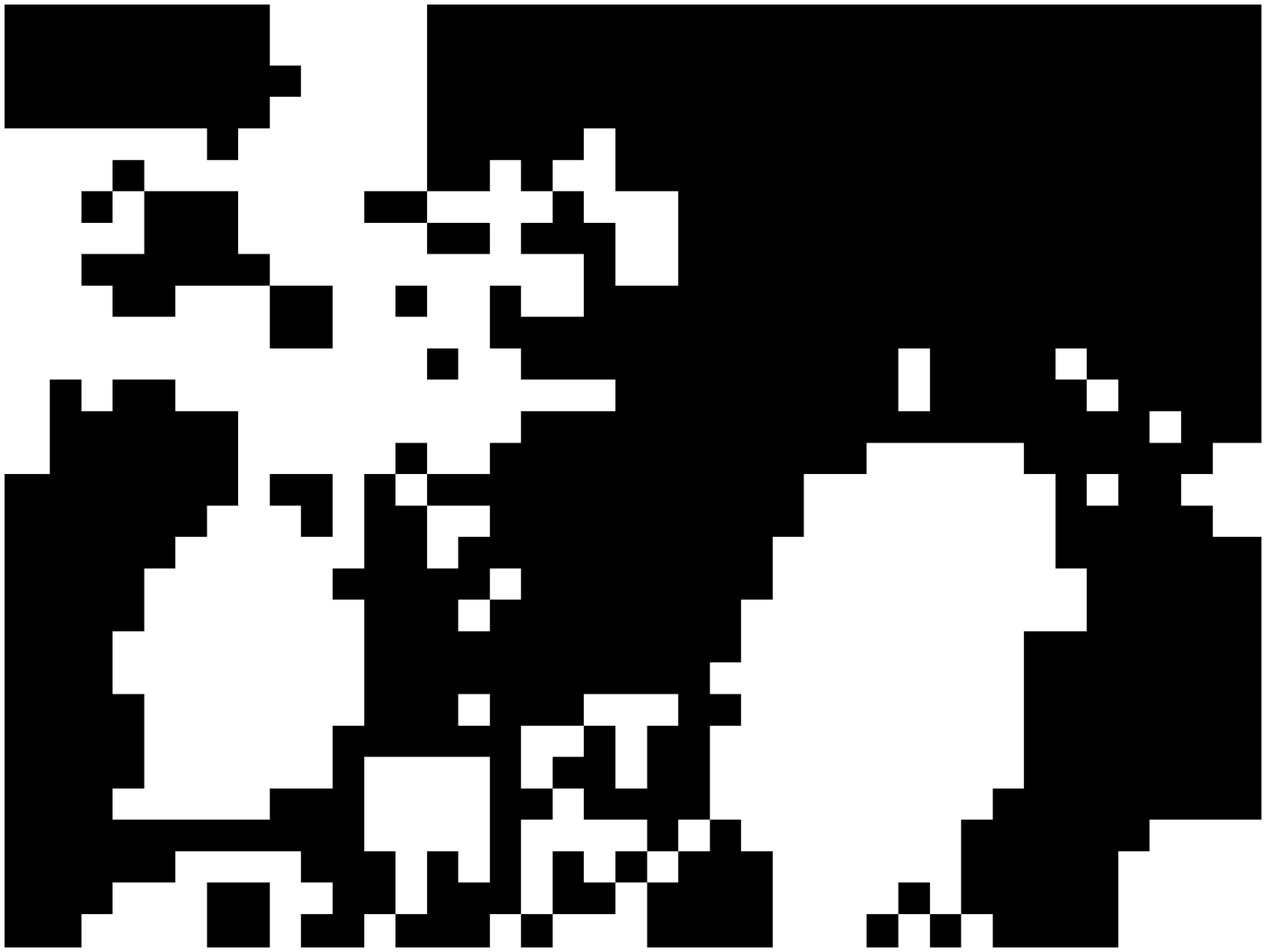}
    \subcaption{(c) Attention mask}
  \end{minipage}
  \caption{Example of attention mask generation.}
  \label{fig:attention_mask}
\end{figure}
}

\newcommand{\figG}{
\begin{figure}[t]
\centering
\includegraphics[width=8cm]{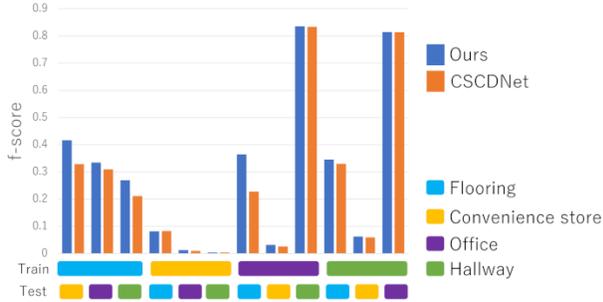}
\caption{Performance results.}\label{fig:Performance}
\end{figure}
}

\newcommand{\figH}{
\begin{figure}[t]
  \centering
  \includegraphics[width=8.5cm]{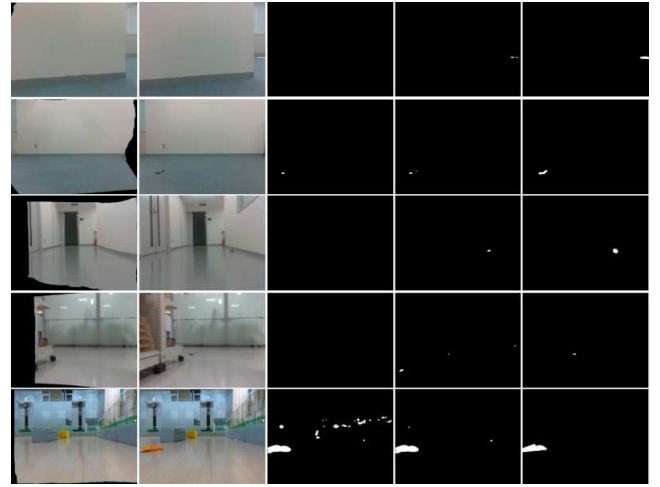}\vspace*{-2mm}
\begin{flushleft}
{\footnotesize
~(a) live image  
(b) ref. image  
(c) CSCDNet  
(d) proposed  
~(e) ground-truth  
}
\end{flushleft}
\caption{Example results.}\label{fig:examples}
\end{figure}
}

\newcommand{\figI}{
\begin{table}[t]
\centering
\caption{Result of ablation study.}
  \begin{tabular}{|r|r|r|r|r|}\hline
                     & conv. store & office & hallway \\ \hline
    Proposed &  0.417 & 0.335 &  0.270  \\ \hline
    W/o PDCNet & 0.063 & 0.350 & 0.247 \\ \hline
    W/o pixel warp &  0.343 & 0.354 & 0.282  \\ \hline
    Reference seq. length: $T=5$ &  0.412 & 0.295 & 0.257  \\ \hline
    Reference seq. length: $T=1$ &  0.356 & 0.310 & 0.233  \\ \hline
  \end{tabular}
\label{Table:ablation_warp_unc}
\end{table}
}

\newcommand{\algA}{
\begin{algorithm}[tb]
\caption{Attention Mask Generation}
\label{alg-attention-mask_gen}
\begin{algorithmic}[1]
  \Require Live image $l_t$ acquired at time $t$, 
reference image sequence $R_t = \{r_i\}_{i = t-T}^{t+T}$ 
acquired within the time-interval $[t-T, t+T]$.
  \Ensure Binary matrix variable $Mask$ with size $W_p \times H_p$ 
  \State Set 1 to each element of $Mask$ 
  \State $LPF <= ExtractPF(l_{t})$
  \ForAll {$r_i \gets R_t$}
  \State $RPF <= ExtractPF(r_{i})$
    \State $MatchMat <= MNN(LPF, RPF)$ 
    \State $Inliers <= RANSAC(MatchMat)$
    \State $Mask <= Binarize(Inliers)  \odot Mask $
  \EndFor
  \State \Return $Mask$
\end{algorithmic}
\end{algorithm}
}

\title{\LARGE \bf
Domain Invariant Siamese Attention Mask 
for Small Object Change Detection 
via Everyday Indoor Robot Navigation
}

\author{Koji Takeda$^{1}$, Kanji Tanaka$^{2}$ and Yoshimasa Nakamura$^{1}$
\thanks{$^{1}$K. Takeda and Y. Nakamura are with Tokyo Metropolitan Industrial Technology Research Institute,
         Tokyo, Japan
        {\tt\small \{takeda.koji\_1, nakamura.yoshimasa\}@iri-tokyo.jp}}%
\thanks{$^{2}$K. Tanaka is with Faculty of Engineering, University of Fukui, Japan.
        {\tt\small tnkknj@u-fukui.ac.jp}}%
}

\maketitle

\begin{abstract}
The problem of image change detection via everyday indoor robot navigation is explored from a novel perspective of the self-attention technique.
Detecting semantically non-distinctive and visually small changes remains a key challenge in the robotics community. 
Intuitively, these small non-distinctive changes may be better handled by the recent paradigm of the attention mechanism, 
which is the basic idea of this work. However, existing self-attention models require significant retraining cost per domain,
so it is not directly applicable to robotics applications. 
We propose a new self-attention technique with an ability of unsupervised on-the-fly domain adaptation, which introduces an attention mask 
into the intermediate layer of an image change detection model, without modifying the input and output layers of the model. 
Experiments, in which an indoor robot aims to detect visually small changes in everyday navigation, 
demonstrate that our attention technique significantly boosts the state-of-the-art image change detection model. 
\end{abstract}

\section{Introduction}

Attention is a technique for selecting a focused location and enhancing different representations of objects at that location. Inspired by the major success of transformer architectures in the field of natural language processing, researchers have recently applied attention techniques to computer vision tasks, such as 
image classification \cite{zhao2020exploring, dosovitskiy2020image}, 
object detection \cite{carion2020end},  
semantic segmentation \cite{xie2021segformer}, 
video understanding \cite{zeng2020learning}, 
image generation \cite{parmar2018image}, 
and pose estimation \cite{lin2021end}. Currently, attention technique is showing it is a potential alternative to CNNs \cite{han2020survey}.

This study explores the attention technique in the context of image change detection for robotics applications.
Image change detection
in 2D perspective views
from an on-board front-facing camera is a fundamental task in robotics 
and has important applications such as novelty detection \cite{contreras2019vision} 
and map maintenance \cite{dymczyk2015gist}.

The problem of 
image change detection becomes 
challenging when 
changes are
semantically 
{\it non-distinctive} 
and visually 
{\it small}.
In these cases, an image change detection model 
(e.g., semantic segmentation 
\cite{sakurada2020weakly}, 
object detection \cite{ObjDetCD}, 
anomaly detection \cite{AnoDetCD}, 
and differencing \cite{alcantarilla2018street}),
which is trained 
in a past domain
to discriminate between the foreground and the background, may fail to classify an unseen object into the correct foreground or background class.
Intuitively, such a small non-distinctive change may be better handled by 
the recent paradigm of
self-attention mechanism, which is the goal of our study.

\figA

Incorporating 
a self-attention mechanism 
into 
an image change detection model
is not straightforward
owing
to the unavailability of 
labeled training data.
Existing attention models 
have primarily been 
studied
in such application domains where rich training data are available \cite{zhao2020exploring}. 
They are typically pre-trained on big data 
and further fine-tuned in the target domain. 
This training process is very expensive for robotics applications,
where robots need to adapt on-the-fly to a new test domain and detect change objects. 
Therefore, a new unsupervised domain-adaptive attention model is required.

We propose a new technique
called domain-invariant attention mask
that
can
adapt
an image change detection model
on-the-fly to
a new target domain,
without modifying the input or output layers,
but
by introducing an attention mechanism to
the intermediate layer (Fig. \ref{fig:tobirae}).
A major advantage of our proposed approach,
owing to
its reliance on
high-level
contextual attention information
rather than low-level visual features,
is its potential to
operate effectively
in test domains
with unseen complex backgrounds.
In this sense,
our approach
combines
the advantages
of two major research directions in the change detection community:
pixel-wise differencing \cite{sakurada2020weakly,chen2021dr}
and
context-based novelty detection \cite{contreras2019vision,pimentel2014review},
by incorporating all available information into the attention mechanism.

Our contributions can be summarized as follows:
(1)
We explore a new approach, 
called domain-adaptive attention model,
to image change detection for robotics applications,
with an ability of unsupervised on-the-fly domain adaptation.
(2)
Instead of considering
pixel-wise differencing \cite{sakurada2020weakly,chen2021dr}
and
context-based novelty detection \cite{contreras2019vision,pimentel2014review},
as two independent approaches,
our framework
combines the advantages
of both approaches
by
incorporating all available
information into the attention mechanism.
(3)
We present a practical system for image change detection
using state-of-the-art techniques such as
image registration \cite{Hausler_2021_CVPR},
pixel warping \cite{truong2021learning},
and
Siamese ConvNet \cite{sakurada2020weakly}.
Experiments,
in which an indoor robot aims to detect visually small changes in everyday navigation,
demonstrate
that our attention technique
significantly boosts
the state-of-the-art image change detection
model.

\section{Related Work}

\subsection{Image Change Detection}

Image change detection is a long standing issue of
computer
vision and it has various applications such as
satellite image \cite{rs11111382,chen2020dasnet}, and autonomous driving \cite{alcantarilla2018street,sakurada2017dense}.
Existing
studies are
divided into 2D or 3D,
according to the sensor modality,
and
we focus on
image change detection
in 2D perspective views
from an on-board front-facing camera
in this study.
Since the camera is a simple and inexpensive sensor, our 2D approach can be expected to have 
an extremely wide range of applications.

Pixel-wise differencing techniques for image change detection rely on the assumption of precise image registration 
between live and reference images \cite{SatelliteCD}.
This method is effective
for 
classical applications
such as satellite imagery \cite{SatelliteCD},
in which
precise registration is available in the form of 2D rotation-translation.
However,
this is not the case for
our
perspective view
applications \cite{PerspectiveCD},
in which
precise pixel-wise registration
itself
is a
challenging
ill-posed problem.
This problem may be alleviated
to some extent
by introducing
an image warping technique,
as we will discuss in Section \ref{sec:pixel_warping}.
However,
such pixel warping is far from perfect,
and may yield false alarms in image change detection.

Novelty detection is a major alternative approach to image change detection \cite{sofman2011anytime}.
In that, novelties are detected as deviations from a nominal image model
that is pre-trained from unlabeled images in a past training domain.
Unlike pixel-wise differencing, 
this technique 
can naturally capture the contextual information of the entire image to determine whether there are any changes in the image.
However,
on the downside,
the change regions cannot be localized within the image
even if the existence of the change is correctly predicted.
Therefore,
existing researches of
novelty detection
in the literature
have focused on
applications
such as 
intruder detection \cite{IntruderDet},
in which
the presence or absence of change, not the position of the changing object, is the most important outcome information.

Several 
new 
architectures
targeting small object change detection have recently been presented.
For example,
Klomp et al. proposed to use Siamese 
CNN to detect markers for improvised explosive devices (IEDs)
\cite{klomp2020real},
where 
they tackled the resolution problem by 
removing the output-side layer of ResNet-18 \cite{he2016deep} to improve the detection performance of small objects.
Our approach differs from these existing approaches
in that
(1)
it does not require to modify the input and output layers of the architecture,
and
(2)
it is able to utilize contextual information.

\subsection{Attention}

Inspired by the major success of transformer architectures in the field of natural language processing, 
researchers have recently applied attention techniques to computer vision tasks, 
such as image classification \cite{zhao2020exploring,dosovitskiy2020image}, 
object detection \cite{carion2020end},  semantic segmentation \cite{xie2021segformer}, video understanding \cite{zeng2020learning}, 
image generation \cite{parmar2018image}, and pose estimation \cite{lin2021end}. 
Because self-attention captures long-range relationships with low computational complexity, 
it is considered a potential alternative to convolutional neural networks (CNNs) \cite{han2020survey}.

Recently, several studies have reported the effectiveness of attention in change detection tasks.
HPCFNet
\cite{HPCFNet}
represents
attention
as 
a correlation between feature maps,
DR-TA Net
\cite{chen2021dr}
evaluates
temporal
attention
by computing
the similarity and dependency
between
a feature map pair,
to realize attention-based change detection.
CSCDNet
\cite{sakurada2020weakly}
employs
a correlation filter
to compensate
for 
the uncertainty
in the non-linear transformation
between live and reference images.

From the perspective of robotic applications,
one of major limitations 
of the current self-attention techniques
is that they require 
a large training set 
to reduce domain dependency. 
In our contribution, 
we introduce a novel 
domain-adaptive 
attention 
technique
that is 
specifically
tailored for 
unsupervised on-the-fly domain adaptation.

\figB

\section{Approach}

Our goal is to incorporate an unsupervised attention model into the image change detection model 
without 
modifying 
the input
and output layers
of the model (Fig. \ref{fig:full}).
In this section,
we 
implement
this idea
on a prototype 
robotic SLAM system.
First,
we perform
a preprocessing 
to compensate for the viewpoint error 
and the resulting uncertainty 
in non-linear mapping from the 3D real environment to 
a 2D image plane of the on-board camera. 
This preprocessing
consists
of 
LRF-SLAM based viewpoint estimation (Section \ref{sec:lrfslam})
followed
by pixel-wise warping 
(Section \ref{sec:pixel_warping}).
However, 
even with 
such a preprocessing,
the images are often affected 
by unpredictable nonlinear mapping errors.
To address this,
we introduce a novel attention mask to direct the robot's attention 
to differentiate the foreground from the background (Section \ref{sec:attention_mask_gen}).
As an advantage,
our approach 
can insert this attention mask into the intermediate layer, 
without modifying the input or output layers (Section \ref{sec:attention_layer}). 
Furthermore, 
we make use of
the pixel-wise confidence 
to further 
improve the image change detection performance (Section \ref{sec:post_processing}). 
The individual modules are detailed as followings.

\figC

\subsection{Dataset Collection}\label{sec:lrfslam}

Figure \ref{fig:mobile_robot}
shows 
the indoor robot experimental platform.
We employ
LRF-SLAM in \cite{lrfslam}
as a method for aligning live images with the reference images.
An input live image 
is paired with 
a reference image if its angle deviation from the live image is less than the threshold 
of 1 degree.
If no such 
reference image exists, 
it 
is paired
with the nearest neighbor viewpoint 
to the live image's viewpoint,
without considering the angle information.

\figD

\subsection{Pixel Warping}

\label{sec:pixel_warping}
We further compensate 
for the viewpoint misalignment in LRF-SLAM by introducing an image warping technique.
A warp is a 2D function, $u(x, y)$, 
which maps a position $(x, y)$ in the reference image to 
a position $u=(x', y')$ in the live image.
Dense image alignment, which is recently proposed in \cite{truong2021learning}, is employed to find an appropriate warp, 
by minimizing an energy function in the form: 
\begin{equation}
-\log p(Y | \Phi(X;\theta))= \sum_{ij}\log p(y_{ij}|\varphi_{ij}(X;\theta))
\end{equation}
where 
$X$ is input image pair $X = (I^q, I^r)$, 
$Y$ is ground-truth flow, $\Phi$ and $\varphi$ are predicted parameters.
An example of pixel warping is shown in Fig. \ref{fig:pixel_warp}. 

\figF

\subsection{Attention Mask Generation}

\label{sec:attention_mask_gen}

We here introduce a novel domain-invariant attention mask (Fig. \ref{fig:attention_mask}), 
inspired by self-attention mechanism \cite{dosovitskiy2020image}.
Recall that
in
standard self-attention \cite{vaswani2017attention},
the interrelationships of the elements in the sequence
are obtained
by computing a weighted sum over all values ${\bf v}$ in the sequence 
for each element in an input sequence ${\bf z} \in R^{N \times D}$. 
The attention weights  are based on the pairwise similarity between
two elements of the sequence and their respective 
query ${\bf q}$ and key ${\bf k}$ representations:
\begin{equation}
  [{\bf q}, {\bf k}, {\bf v}] = {\bf zU}_{qkv}  \hspace{2cm}   {\bf U}_{qkv} \in \mathbb{R}^{D \times 3D_h}
\end{equation}

\begin{equation}
  \label{eq:sa}
SA({\bf z}) = softmax({\bf qk}^T / \sqrt{D_h}){\bf v} 
\end{equation}

In the proposed method, 
this {\it SA} term is replaced with:
\begin{equation}
Proposed({\bf q_{p}},{\bf k_{p}},{\bf m_{cnn}}) = PatchMatch({\bf q_{p}},{\bf k_{p}}) \odot {\bf m_{cnn}}.
\end{equation}
Here, ${\bf q_{p}} \in \mathbb{R}^{h_{np} \times w_{np} \times D_p}$ 
and 
${\bf k_{p}} \in \mathbb{R}^{h_{np} \times w_{np} \times D_p}$ 
are patches extracted from live and reference images, respectively. 
${\bf m_{cnn}} \in \mathbb{R}^{h_{cnn} \times w_{cnn} \times D_{cnn}}$
is an intermediate feature of the Siamese CNN. The $PatchMatch$ is 
the function 
that predicts whether 
or not a pair of $D_p$-dim vectors of ${\bf q_{p}}$ and ${\bf k_{p}}$ match.

We generate a binary attention mask by incorporating the attention mechanism.
First,
the image is reshaped into a sequence of 2D patches, each of which is described by a local feature vector.
We employ the 128-dim deep PatchNetVLAD \cite{Hausler_2021_CVPR}
descriptor as the local feature vector.
The attention score is then computed for each region of interest. 
We 
then 
evaluate
the attention score
as the 
patch-wise 
dissimilarity (i.e., L2 distance) 
between 
live and reference image pairs.
Then,
RANSAC geometric verification is performed to filter out
false alarms
that are 
originated from change patches.
Finally,
we 
obtain
the attention regions
as 
scattered discrete regions of
live patches
with positive attention score,
which
makes a binary attention mask.

\algA

Algorithm \ref{alg-attention-mask_gen} presents the algorithm for creating the attention mask.
It aims to 
compute 
a binary 
attention mask 
$Mask$
for 
an array of
$W_p\times H_p$
patches
at time instance $t$
from
a sequence of live images within 
the time interval $[t-T, t+T]$.
The algorithm
begins with 
the initialization of the mask variable $Mask$,
and 
iterates for each live image,
the following steps:
First,
it extracts 
from an input image
a set of 
PatchNetVLAD feature vectors (``$ExtractPF$''),
each of which 
belongs to one of reference patches.
Then,
for each live feature, 
it searches
for
its mutual nearest neighbor 
(``$MNN$")
reference patch 
in terms of the L2 norm of their PatchNetVLAD features.
Here,
the mutual nearest neighbor search is defined as the process of searching for pairs of matching live and reference elements that 
are closest to each other.
Only feature pairs that have passed the mutual nearest neighbor search are sent to the next RANSAC process.
Then,
it performs 
geometric verification by RANSAC \cite{ransac} (``$RANSAC$").
Finally,
it outputs
pixel
with
values greater than or equal to threshold,
in the form of the binary attention mask:
\begin{equation}
\label{eq:binary_elem}
\mathbf{b}[i,j] = \left\{
\begin{array}{ll}
1 & \mbox{If $inliers[i,j]$ passed RANSAC}\\
0 & \mbox{Otherwise}
\end{array}
\right.
.
\end{equation}

\subsection{Attention Mask Layer}\label{sec:attention_layer}

We 
now 
insert 
the attention mask 
into the standard
image change detection model 
of the Siamese CNN (Section \ref{sec:attention_mask_gen}). 
For the Siamese CNN, we use the state-of-the-art architecture of CSCDNet \cite{sakurada2020weakly}. 
The attention mask layer takes the CNN feature map and attention mask as inputs and outputs the CNN features masked in the channel direction. 
We inserted the attention mask before correlation operation (i.e., before concatenating decoded feature).

We perform the process of masking the CNN Siamese feature map in the channel direction.
Let
$\mathbf{fmap_{new}} \in \mathbf{R}^{W \times H}$ 
denote the feature map after attention is applied. 
Let
$\mathbf{fmap_{old}} \in {R}^{W \times H \times C}$ 
denote the feature map obtained from Siamese CNN. 
Here, $W$ denote the tensor width, 
$H$ denote the tensor height, 
and $C$ denote the tensor channel. 
Let
$\mathbf{mask} \in \mathbf{R}^{W \times H}$ 
denote the attention mask. 
Then,
the attention mask element 
at the $i$-th row,
$j$-th column
and $k$-th channel is:
\begin{equation}
\label{eq:merge}
  \mathbf{fmap_{new}}[i,j,k] = \mathbf{fmap_{old}}[i,j,k] \cdot \mathbf{mask}[i,j]. 
\end{equation}
This operation is applied to the both branches of the Siamese CNN.

\subsection{Post Processing}\label{sec:post_processing}

Post-processing is introduced to eliminate false alarms in the detection results.
We evaluate
the uncertainty 
in the output layer of 
the dense image alignment model
and use it to evaluate
the confidence of prediction at each pixel.
Intuitively,
a high probability of pixel warping uncertainty 
indicates 
that no corresponding pixel exists; 
therefore is a high possibility of change. 
Conversely, 
low probability
indicates that the corresponding pixel exists; 
therefore is a low possibility of change. 
This masking process can be simply expressed as 
an 
Hadamard product
operation, in the following form:
\begin{equation}
\label{eq:unc_merge}
\mathbf{output_{new}}[i,j] = \mathbf{output_{old}}[i,j] \cdot \mathbf{uncertainty}[i,j].
\end{equation}
Here, $\mathbf{output_{old}} \in \mathbf{R}^{W \times H}$ represents the change probability value map of the output of 
the Siamese CNN. 
$\mathbf{uncertainty} \in \mathbf{R}^{W \times H}$ represents the uncertainty of each pixel warp of a live image. 
$\mathbf{output_{new}} \in \mathbf{R}^{W \times H}$  represents the probability of change for each pixel after the merging process.

\section{Evaluation Experiments}

\subsection{Dataset}

We collected 
four datasets,
``convenience store,''
``flooring,''
``office room,''
and ``hallway,''
in four distinctive environments.
Eight independent image sequences 
are collected
from the four different environments.
The number of images
are
534, 491, 378, and 395,
respectively
for
``flooring," 
``convenience store," 
``office," 
and
``hallway".
Examples of these images are shown in Fig. \ref{fig:dataset}.
The image size was $640\times 480$. 
The 
ground-truth change object regions 
in each live image
are manually annotated
using PaddleSeg \cite{liu2021paddleseg, paddleseg2019}
as the annotation tool.

\figE

\subsection{Settings}

The state-of-the-art model,
CSCDNet \cite{sakurada2020weakly},
is used as our base architecture,
which we aim to boost in this study.
It is also used as a comparing method to verify 
whether the proposed method 
can actually boost the CSCDNet.
The network is initialized 
with the weight pre-trained on ImageNet \cite{deng2009imagenet}.
The pixel-wise binary cross-entropy loss is used as loss function as in the original work of CSCDNet \cite{sakurada2020weakly}. 
PDCNet \cite{truong2021learning}
is used to align reference images.
Adam optimizer \cite{kingma2014adam} 
is used for the network training.
Learning rate is 0.0001. 
The number of iterations is 20,000.
The batch size is 32.
A nearest neighbor interpolation
is used 
to resize the attention mask
to 
fit into the attention mask layer. 
The length of reference image sequence is set to $T=10$ in default.
A single NVIDIA GeForce RTX 3090 GPU 
with PyTorch framework is used. 
Pixel-wise precision, recall, and F1 score are used as performance index.

\figG

\subsection{Quantitative Results}

Figure \ref{fig:Performance}
shows performance results. 
As can be seen,
the proposed method 
outperformed the comparing method
for almost all combinations 
of training and test sets
considered here.
Notably, the proposed method 
extremely
outperformed the comparing method when it was trained on the ``flooring'' dataset. 
The ``flooring'' is the simplest background scene. 
Therefore, the image change detection model trained on that could be generalized to other complex domains as well. 
However, the proposed method performed almost the same as the comparing method when it was trained on the other complex background scenes. 
As an example, for the convenience store dataset, 
the robot navigates through a narrow and messy passage that makes its visual appearance very different from that of the other two datasets.
This makes 
the 
proposed
training
algorithm
less effective for 
the training set.
It is noteworthy
that 
such an effect
of visual appearance 
might be 
mitigated by introducing
view synthesis technique such as \cite{kupyn2019deblurgan},
which is a direction of future research.

\figH

\subsection{Qualitative Results}

Figure \ref{fig:examples} shows example results
in which 
the proposed attention mechanism
was 
often
successful 
in improving the performance of the CSCDNet.
Especially,
the proposed method
was effective for 
complex background scenes,
owing to the ability of the proposed attention mechanism
to make use of contextual information.

\figI

\subsection{Ablation Study}

Table \ref{Table:ablation_warp_unc}
presents the results of 
a series of ablation studies 
by turning off some of the modules in the proposed framework.
For all ablations, the flooring dataset is used as the training data.
For the ``convenience store'' dataset,
the performance is 
significantly higher
with than without 
the post-processing technique
in Section \ref{sec:post_processing}.
Because of the complex background of this dataset, 
pixel warping often fails, 
and the proposed method was effective in suppressing 
such effects
that are originated from complex backgrounds.
For the ``office'' and ``hallway'' datasets,
the performance is almost the same 
between
with and without
the technique.
Since the background was less complex in these datasets, 
pixel warp failures were less common, 
therefore the effect of estimating uncertainty was small. 
Next, another ablation with different settings of the length of reference image sequences are conducted. 
As can be seen, the performance is best at $T$=10.
As expected, higher performance was obtained with longer reference sequences.
From the above results,
it could be concluded that 
both of PDCNet and pixel warping 
play important roles
and
can actually improve the performance
of image change detection.

\section{Conclusions}

In this research, 
we tackled the challenging problem of small object change 
detection 
via everyday indoor robot navigation.
We proposed
a new self-attention technique with 
unsupervised on-the-fly domain adaptation,
by introducing an attention mask into the intermediate layer of 
an image change detection model,
without modifying the input and output layers
of the model.
Experiments using a novel dataset on small object change detection verified 
that
the proposed method significantly boosted the state-of-the-art model for image change detection.

\bibliography{reference} 
\bibliographystyle{unsrt}

\end{document}